\documentclass{article}
\usepackage{comment}

\title{Light Field Networks: Neural Scene Representations with Single-Evaluation Rendering}

\author{%
    Vincent Sitzmann\textsuperscript{1,}\thanks{These authors contributed equally to this work.} \\
	\texttt{sitzmann@mit.edu} \\
	\And
	Semon Rezchikov\textsuperscript{2,}\footnotemark[1]\\
	\texttt{skr@math.columbia.edu}
	\AND
	William T. Freeman\textsuperscript{1,3}\\
	\texttt{billf@mit.edu}
	\And
	Joshua B. Tenenbaum\textsuperscript{1,4,5}\\
	\texttt{jbt@mit.edu}
	\And
	Fr\'{e}do Durand\textsuperscript{1}\\
	\texttt{fredo@mit.edu}
}

\usepackage[nonatbib,final]{neurips_2021}
\usepackage[numbers,compress]{natbib}

\usepackage[utf8]{inputenc} %
\usepackage[T1]{fontenc}    %
\usepackage{hyperref}       %
\usepackage{url}            %
\usepackage{booktabs}       %
\usepackage{amsfonts}       %
\usepackage{amsmath}
\usepackage{amssymb}
\usepackage{nicefrac}       %
\usepackage{microtype}      %
\usepackage{hyperref}
\usepackage[dvipsnames]{xcolor}
\usepackage{comment}

\hypersetup{
    colorlinks=true,
    linkcolor=MidnightBlue,
    citecolor=black
}
\usepackage{graphicx} 
\usepackage{caption,subcaption}
\usepackage{overpic}
\usepackage{amsmath}
\usepackage{physics}
\usepackage{color}
\usepackage{wrapfig}
\usepackage{cleveref}
\usepackage{colortbl}
\usepackage{textcomp}

\usepackage{booktabs} %
\usepackage{multirow}
\usepackage{subcaption}
\usepackage{tabularx}
\usepackage{amsthm}
\usepackage{pdfpages}

\theoremstyle{definition}
\newtheorem{proposition}{Proposition}

\usepackage{soul}
\usepackage{color}
\usepackage{xcolor}
\definecolor{turquoise}{cmyk}{0.65,0,0.1,0.3}
\definecolor{purple}{rgb}{0.65,0,0.65}
\definecolor{dark_green}{rgb}{0, 0.5, 0}
\definecolor{orange}{rgb}{0.8, 0.6, 0.2}
\definecolor{red}{rgb}{0.8, 0.2, 0.2}
\definecolor{darkred}{rgb}{0.6, 0.1, 0.05}
\definecolor{blueish}{rgb}{0.0, 0.3, .6}
\definecolor{light_gray}{rgb}{0.7, 0.7, .7}
\definecolor{pink}{rgb}{1, 0, 1}
\definecolor{greyblue}{rgb}{0.25, 0.25, 1}
\definecolor{tab_blue}{HTML}{1f77b4}
\definecolor{tab_orange}{HTML}{ff7f0e}
\definecolor{LightRed}{rgb}{0.99,0.89,0.89}
\definecolor{mesh_misty_rose}{HTML}{e6aaa3}
\definecolor{mesh_yellow}{HTML}{ffba00}

\newcommand{\ours}{Light Field Network}
\newcommand{\ourshort}{LFN}
\newcommand{\plucker}{Pl{\"u}cker}

\newcommand{\LF}{\Phi}

\newcommand{\ray}{\mathbf{r}}
\newcommand{\point}{\mathbf{p}}
\newcommand{\dir}{\mathbf{d}}
\newcommand{\moment}{\mathbf{m}}
\newcommand{\render}{\mathbf{m}}
\newcommand{\threedrep}{\Phi^{3D}}
\newcommand{\R}{\mathbb{R}}
\newcommand{\dataset}{\mathcal{D}}
\newcommand{\image}{\mathbf{I}}
\newcommand{\intrinsiccamera}{\mathbf{K}}
\newcommand{\extrinsiccamera}{\mathbf{E}}
\newcommand{\hypernetwork}{\Psi}
\newcommand{\latent}{\mathbf{z}}
\newcommand{\lfparameter}{\phi}
\newcommand{\hypernetworkparameter}{\psi}
\newcommand{\renderingmap}{\Theta}
\newcommand{\ourc}{\mathbf{c}}

\newcommand{\linea}{\mathbf{a}}
\newcommand{\lineb}{\mathbf{b}}

\DeclareMathOperator*{\argmin}{arg\,min}

\definecolor{MyDarkBlue}{rgb}{0.02,0.02,0.6}

\begin{document}

\maketitle

\vbox{%
	\vskip -0.26in
	\hsize\textwidth
	\linewidth\hsize
	\centering
	\normalsize
	\footnotemark[1]MIT CSAIL \quad \footnotemark[2]Columbia University \quad \footnotemark[3] IAFI \quad \footnotemark[4]MIT BCS \quad \footnotemark[5] CBMM\\
	\tt\href{https://vsitzmann.github.io/lfns/}{vsitzmann.github.io/lfns/}
	\vskip 0.3in
}

\begin{abstract}
    Inferring representations of 3D scenes from 2D observations is a fundamental problem of computer graphics, computer vision, and artificial intelligence.
    Emerging 3D-structured neural scene representations are a promising approach to 3D scene understanding.
    In this work, we propose a novel neural scene representation, \ours{}s or \ourshort{}s, which represent both geometry and appearance of the underlying 3D scene in a 360-degree, four-dimensional light field parameterized via a neural network. 
    Rendering a ray from an \ourshort{} requires only a \emph{single} network evaluation, as opposed to hundreds of evaluations per ray for ray-marching or volumetric based renderers in 3D-structured neural scene representations. 
    In the setting of simple scenes, we leverage meta-learning to learn a prior over \ourshort{}s that enables multi-view consistent light field reconstruction from as little as a single image observation.
    This results in dramatic reductions in time and memory complexity, and enables real-time rendering.
    The cost of storing a 360-degree light field via an \ourshort{} is two orders of magnitude lower than conventional methods such as the Lumigraph. 
    Utilizing the analytical differentiability of neural implicit representations and a novel parameterization of light space, we further demonstrate the extraction of sparse depth maps from \ourshort{}s. 
\end{abstract}

\section{Introduction}
A fundamental problem across computer graphics, computer vision, and artificial intelligence is to infer a representation of a scene's 3D shape and appearance given impoverished observations such as 2D images of the scene.
Recent contributions have advanced the state of the art for this problem significantly. 
First, neural implicit representations have enabled efficient representation of local 3D scene properties by mapping a 3D coordinate to local properties of the 3D scene at that coordinate~\cite{park2019deepsdf,mescheder2019occupancy,sitzmann2019srns,mildenhall2020nerf,Niemeyer2020DVR,yu2020pixelnerf}.
Second, differentiable neural renderers allow for the inference of these representations given only 2D image observations~\cite{sitzmann2019srns,mildenhall2020nerf}.
Finally, leveraging meta-learning approaches such as hypernetworks or gradient-based meta-learning has enabled the learning of distributions of 3D scenes, and therefore reconstruction given only a single image observation~\cite{sitzmann2019srns}.
This has enabled a number of applications, such as novel view synthesis~\cite{mildenhall2019local,sitzmann2019srns,yu2020pixelnerf}, 3D reconstruction~\cite{Niemeyer2020DVR,sitzmann2019srns} semantic segmentation~\cite{zhi2021place,kohli2020semantic}, and SLAM~\cite{sucar2021imap}.
However, 3D-structured neural scene representations come with a major limitation: Their rendering is prohibitively expensive, on the order of \emph{tens of seconds for a single $256\times256$ image} for state-of-the-art approaches. In particular,  parameterizing the scene in 3D space necessitates the discovery of surfaces along camera rays during rendering.
This can be solved either by encoding geometry as a level set of an occupancy or signed distance function, or via volumetric rendering, which solves an alpha-compositing problem along each ray.
Either approach, however, requires tens or even hundreds of evaluations of the 3D neural scene representation in order to render a single camera ray. 

We propose a novel neural scene representation, dubbed \ours{}s or \ourshort{}s. Instead of encoding a scene in 3D space, \ours{}s encode a scene by directly mapping an oriented camera ray in the four dimensional space of light rays to the radiance observed by that ray. 
This obviates the need to query opacity and RGB at 3D locations along a ray or to ray-march towards the level set of a signed distance function, speeding up rendering by \emph{three orders of magnitude} compared to volumetric methods.
In addition to directly encoding appearance, we demonstrate that \ourshort{}s encode information about scene geometry in their derivatives.
Utilizing the unique flexibility of neural field representations, we introduce the use of \plucker{} coordinates to parameterize 360-degree light fields, which allow for storage of a-priori unbounded scenes and admit a simple expression for the depth as an analytical function of an \ourshort{}.
Using this relationship, we demonstrate the computation of geometry in the form of sparse depth maps.
While 3D-structured neural scene representations are multi-view consistent \emph{by design}, parameterizing a scene in light space does not come with this guarantee: the additional degree of freedom enables rays that view the same 3D point to change appearance across viewpoints.
For the setting of simple scenes, we demonstrate that this challenge can be overcome by learning a prior over 4D light fields in a meta-learning framework. 
We benchmark with current state-of-the-art approaches for single-shot novel view synthesis, and demonstrate that \ourshort{}s compare favorably with globally conditioned 3D-structured representations, while accelerating rendering and reducing memory consumption by orders of magnitude.

In summary, we make the following contributions:
\begin{enumerate}
    \item We propose \ours{}s (\ourshort{}s), a novel neural scene representation that directly parameterizes the light field of a 3D scene via a neural network, enabling real-time rendering and vast reduction in memory utilization. 
    \item We demonstrate that we may leverage 6-dimensional \plucker{} coordinates as a parameterization of light fields, despite their apparent overparameterization of the 4D space of rays, thereby enabling continuous, 360-degree light fields.
    \item By embedding \ourshort{}s in a meta-learning framework, we demonstrate light field reconstruction and novel view synthesis of simple scenes from sparse 2D image supervision only.
    \item We demonstrate that inferred \ourshort{}s encode both appearance and geometry of the underlying 3D scenes by extracting sparse depth maps from the derivatives of \ourshort{}s, leveraging their analytical differentiability.
\end{enumerate}
\vspace{-5pt}

\paragraph{Scope.} The proposed method is currently constrained to the reconstruction of simple scenes, such as single objects and simple room-scale scenes, in line with recent work on learning generative models in this regime~\cite{sitzmann2019srns,kosiorek2021nerf}.
\section{Related Work}

\paragraph{Neural Scene Representations and Neural Rendering.}
A large body of work addresses the question of inferring feature representations of 3D scenes useful to downstream tasks across graphics, vision, and machine learning.
Models without 3D structure suffer from poor data efficiency~\cite{worrall2017interpretable,eslami2018neural}.
Voxel grids~\cite{sitzmann2018deepvoxels,tung2018learning, phuoc2018rendernet,zhu2018visual,Phuoc2019HoloGAN,lombardi2019neural,Rezende2016unsupervised} offer 3D structure, but scale poorly with spatial resolution.
Inspired by neural implicit representations of 3D geometry~\cite{park2019deepsdf,mescheder2019occupancy}, recent work has proposed to encode properties of 3D scenes as \emph{neural fields} (also implicit- or coordinate-based representations, see ~\cite{xie2021neuralfield} for an overview), neural networks that map 3D coordinates to local properties of the 3D scene at these coordinates. 
Using differentiable rendering, these models can be learned from image observations only~\cite{sitzmann2019srns,mildenhall2020nerf,tewari2020state,kosiorek2021nerf}.
Reconstruction from sparse observations can be achieved by learning priors over the space of neural fields~\cite{sitzmann2019srns,Niemeyer2020DVR,kosiorek2021nerf,sitzmann2020metasdf,sitzmann2019siren,RematasICML21} or by conditioning of the neural field on local features~\cite{yu2020pixelnerf,saito2019pifu,trevithick2020grf}.
Differentiable rendering of such 3D-structured neural scene representations is exceptionally computationally intensive, requiring hundreds of evaluations of the neural representation per ray, with tens of thousands to millions of rays per image.
Some recent work seeks to accelerate test-time rendering, but either does not admit generalization~\cite{reiser2021kilonerf,yu2021plenoctrees,liu2020neural}, or does not alleviate the cost of rendering at training/inference time ~\cite{lindell2020autoint,neff2021donerf,kellnhofer2021neural}.
With \ours{}s, we propose to leverage 360-degree light fields as neural scene representations. We introduce a novel neural field parameterization of 360-degree light fields, infer light fields  via meta-learning from as few as a single 2D image observation, and demonstrate that \ourshort{}s encode both scene geometry and appearance.

\paragraph{Light fields and their reconstruction.}
Light fields have a rich history as a scene representation in both computer vision and computer graphics.
Adelson et al.~\cite{adelson1991plenoptic} introduced the 5D plenoptic function as a unified representation of information in the early visual system~\cite{adelson1992single}.
Levoy et al.~\cite{levoy1996light} and, concurrently, Gortler et al.~\cite{gortler1996lumigraph} introduced light fields in computer graphics as a 4D sampled scene representation for fast image-based rendering.
Light fields have since enjoyed popularity as a representation for novel view synthesis~\cite{buehler2001unstructured} and computational photography, e.g.~\cite{ng2006digital}.
Light fields enable direct rendering of novel views by simply extracting a 2D slice of the 4D light field. However, they tend to incur significant storage cost, and since they rely on two-plane parameterizations, they make it hard to achieve a full 360-degree representation without concatenating multiple light fields.
A significant amount of prior work addresses reconstruction of fronto-parallel light fields via hand-crafted priors, such as sparsity in the Fourier or shearlet domains~\cite{vagharshakyan2017light,shi2014light,levin2010linear}.
With the advent of deep learning, approaches to light field reconstruction that leverage convolutional neural networks to in-paint or extrapolate light fields from sparse views have been proposed~\cite{kalantari2016learning,mildenhall2019local,Bemana2020xfields}, but similarly only support fronto-parallel novel view synthesis.
We are instead interested in light fields as a representation of 3D appearance and geometry that enables efficient inference of and reasoning about the properties of the full underlying scene.
\begin{figure*}
	\centering
	\includegraphics[width=\textwidth]{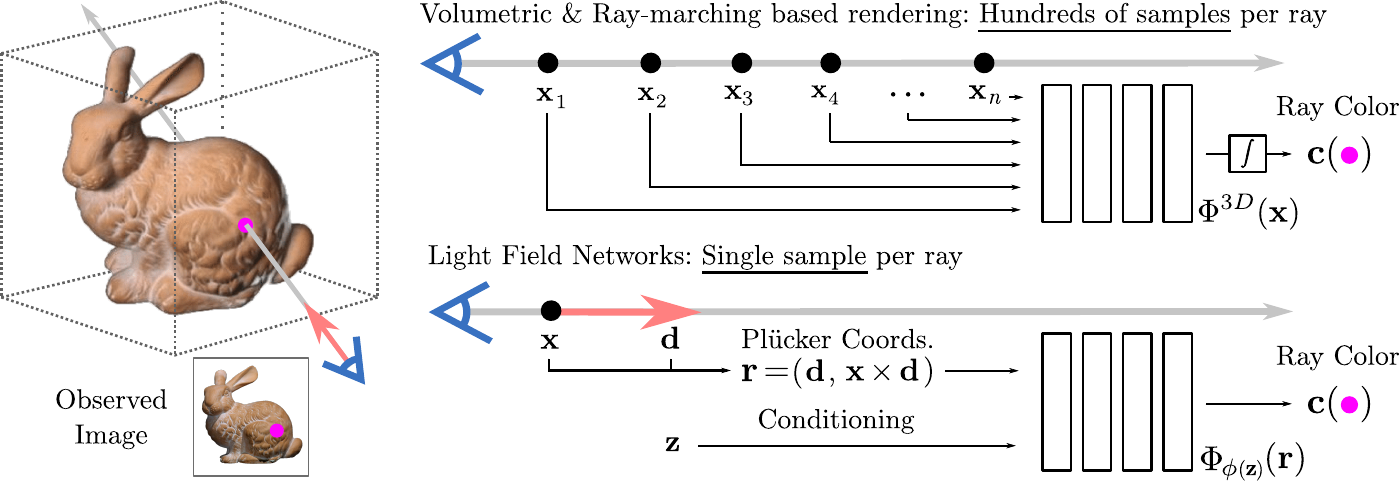}
	\vspace{-0.4cm}
	\caption{\textbf{Overview.} We propose \ours{}s or \ourshort{}s, which encode the full 360-degree light field of a 3D scene in the weights of a fully connected neural network $\LF_\phi$ (with weights $\phi$ conditioned on a latent code $\latent$) that maps an oriented ray $\mathbf{r}$ to the radiance $\mathbf{c}$ observed by that ray.
    Rendering an \ourshort{} $\LF_\phi$ only requires evaluating the underlying MLP once per ray, in contrast to 3D-structured neural scene representations $\Phi_{3D}$ such as SRNs~\cite{sitzmann2019srns}, NeRF~\cite{mildenhall2020nerf}, or DVR~\cite{Niemeyer2020DVR} that require hundreds of evaluations per ray.
	We leverage meta-learning to learn a multi-view consistent space of \ourshort{}s.
	Once trained, this enables reconstruction of a 360-degree light field and subsequent real-time novel view synthesis of simple scenes from only a single observation.}
	\label{fig:overview}
	\vspace{-0.32cm}
\end{figure*}

\section{Background: 3D-structured Neural Scene Representations}
\label{sec:3D_nsr}
Recent progress in neural scene representation and rendering has been driven by two key innovations.
The first are neural fields, often also referred to as neural implicit- or coordinate-based scene representations~$\threedrep$~\cite{sitzmann2019srns,mildenhall2020nerf}, which model a scene as a continuous function, parameterized as an MLP which maps a 3D coordinate to a representation $\mathbf{v}$ of whatever is at that 3D coordinate:
\begin{equation}
    \label{eq:threedrep}
    \threedrep: \mathbb{R}^3 \to \mathbb{R}^n, \quad \mathbf{x} \mapsto \threedrep(\mathbf{x}) = \mathbf{v}.
\end{equation}
The second is a differentiable renderer $\render$, which, given a ray $\ray$ in $\R^3$, and the representation $\threedrep$, computes the value of the color $\mathbf{c}$ of the scene when viewed along $\ray$:
\begin{equation}
    \label{eq:threedrender}
    \render(\ray, \threedrep) = \mathbf{c}(\ray) \in \mathbb{R}^3. 
\end{equation}

Existing rendering methods broadly fall into two categories: sphere-tracing-based renderers~\cite{sitzmann2019srns,yariv2020multiview,Niemeyer2020DVR,DISN} and volumetric renderers \cite{lombardi2019neural,mildenhall2020nerf}. These  methods require on the order of tens or hundreds of evaluations of the values of $\threedrep$ along a ray $\ray$ to compute $\mathbf{c}(\ray)$. 
This leads to extraordinarily large memory and time complexity of rendering. As training requires error backpropagation through the renderer, this impacts both training and test time.

\section{The \ours{} Scene Representation}
We propose to represent a scene as a 360-degree \emph{neural light field}, a function parameterized by an MLP $\Phi_\phi$ with parameters $\phi$ that directly maps the 4D space $\mathcal{L}$ of oriented rays to their observed radiance:
\begin{equation}
    \LF_\phi: \mathcal{L} \to \mathbb{R}^3, \ray \mapsto \Phi_\phi(\ray) = \mathbf{c}(\ray). 
\end{equation}
A light field completely characterizes the flow of light through unobstructed space in a static scene with fixed illumination.
Light fields have the unique property that rendering is achieved by a \emph{single evaluation} of $\LF$ per light ray, i.e., \emph{no ray-casting is required}. 
Moreover, while the light field only encodes appearance explicitly, its derivatives encode geometry information about the underlying 3D scene~\cite{bolles1987epipolar,adelson1991plenoptic,adelson1992single}.
This makes many methods to extract 3D geometry from light fields possible \cite{dansereau2004gradient,tosic2014light, kim2013scene, si2016dense}, and we demonstrate efficient recovery of sparse depth maps from \ourshort{}s below.  

\subsection{Implicit representations for 360 degree light fields}
To fully represent a 3D scene requires a parameterization of all light rays in space. Conventional light field methods are constrained to leverage minimal parameterizations of the 4D space of rays, due to the high memory requirements of discretely sampled high-dimensional spaces. 
In contrast, our use of neural field representations allows us to freely choose a continuous parameterization that is mathematically convenient.
In particular, we propose to leverage the 6D \plucker{} parameterization of the space of light rays $\mathcal{L}$ for \ourshort{}s. 
The \plucker{} coordinates (see~\cite{jia2020plucker} for an excellent overview) of a ray $\ray$ through a point $\point$ in a normalized direction $\dir$ are  
\begin{equation}
    \label{eq:plucker}
    \ray = (\dir, \moment) \in \R^6 \text{ where }  \mathbf{m} = \point \times \dir, \text{ for } \quad \dir \in \mathbb{S}^2, \point \in \mathbb{R}^3.
\end{equation}
where $\times$ denotes the cross product.
While \plucker{} coordinates are a-priori $6$-tuples of real numbers, the coordinates of any ray lie on a curved 4-dimensional subspace $\mathcal{L}$. 
\plucker{} coordinates uniformly represent all oriented rays in space without singular directions or special cases. Intuitively, a general ray $\ray$ together with the origin define a plane, and $\mathbf{m}$ is a normal vector to the plane with its magnitude capturing the distance from the ray to the origin; if $\mathbf{m}=0$ then the ray passes through the origin and is defined by its direction $\mathbf{d}$. 
This is in contrast to conventional light field parameterizations:
Fronto-parallel two-plane or cylindrical parameterizations cannot represent the full 360-degree light field of a scene~\cite{levoy1996light,krolla2014spherical}.
Cubical two-plane arrangements~\cite{gortler1996lumigraph,buehler2001unstructured} are not continuous, complicating the parameterization via a neural implicit representation.
In contrast to the two-sphere parameterization~\cite{ihm1997spherical}, \plucker{} coordinates do not require that scenes are bounded in size and do not require spherical trigonometry.

The parameterization via a neural field enables compact storage of a 4D light field that can be sampled at arbitrary resolutions, while non-neural representations are resolution-limited.
Neural fields further allow the analytical computation of derivatives. This enables the efficient computation of sparse depth maps, where prior representations of light fields require finite-differences approximations of the gradient~\cite{dansereau2004gradient,tosic2014light, kim2013scene}.

\paragraph{Rendering \ourshort{}s.}
To render an image given an \ourshort{}, one computes the \plucker{} coordinates $\ray_{u,v}$ of the camera rays at each $u$, $v$ pixel coordinate in the image according to \autoref{eq:plucker}.
Specifically, given the extrinsic $\mathbf{E} = \big[\mathbf{R}|\mathbf{t}\big] \in SE(3)$ and intrinsic $\mathbf{K}\in \mathbb{R}^{3\times3}$ camera matrices~\cite{Hartley:2003} of a camera, one may retrieve the \plucker{} coordinates of the ray $\ray_{u,v}$ at pixel coordinate $u$, $v$ as: 
\begin{equation}
\ray_{u,v} = (\dir_{u,v}, \mathbf{t} \times \dir_{u,v}) / \|\dir_{u,v}\|, \text{ where }
    \dir_{u,v} = \mathbf{R}\mathbf{K}^{-1}
    \left(
    \begin{smallmatrix}
    u\\
    v\\
    1\\
    \end{smallmatrix}
    \right)
    + \mathbf{t},
\end{equation}
where we use the world-to-camera convention for the extrinsic camera parameters.
Rendering then amounts to a \emph{single} evaluation of the \ourshort{} $\LF$ for each ray, $\mathbf{c}_{u,v}=\LF(\ray_{u,v})$. For notational convenience, we introduce a rendering function 
\begin{equation}
\label{eq:renderer}
    \renderingmap^\LF_{\extrinsiccamera, \intrinsiccamera}:\R^\ell  \to \R^{H \times W \times 3}
\end{equation} 
which renders an \ourshort{} $\LF_\phi$ with parameters $\lfparameter \in \R^\ell$ when viewed from a camera with extrinsic and intrinsic parameters $(\extrinsiccamera, \intrinsiccamera)$ into an image. 

\subsection{The geometry of \ours{}s}
\label{sec:structure}
\begin{figure*}
	\centering
    \includegraphics[width=\textwidth]{"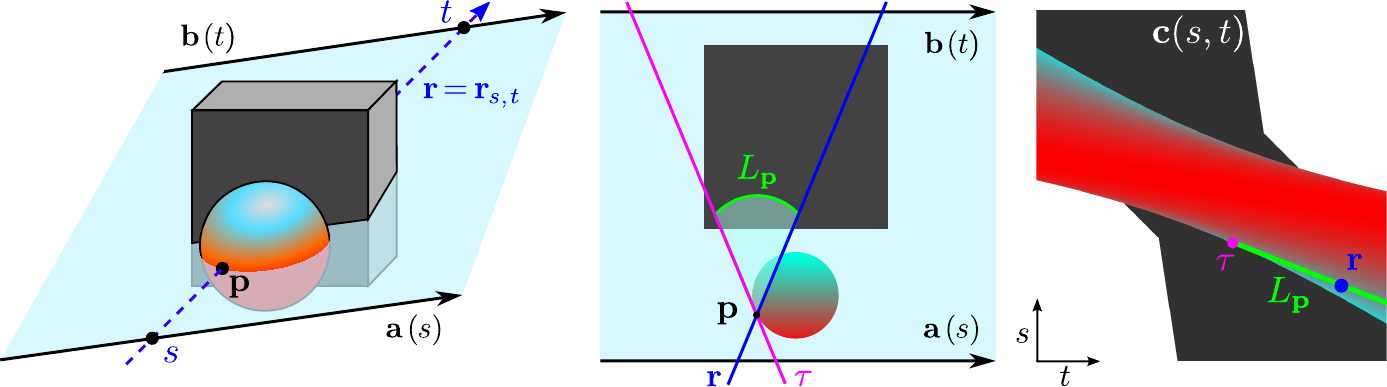"}
	\caption{Given a 3D scene (left) and a light ray $\mathbf{r}$ (blue), we can slice the scene along a 2D plane containing the ray (light blue), yielding a 2D scene (center).
	The light field of all rays in a 2D plane, the \emph{Epipolar Plane Image} (EPI) $\mathbf{c}(s, t)$ (right).
	can be analytically computed from our 360-degree LFN. 
	The family of rays $L_\mathbf{p}$ (green) going through a point $\mathbf{p}$ on an object in the scene defines a straight line in the EPI.  See below for further discussion of EPI geometry.
	}
	\label{fig:2D_light_field}
\end{figure*}

We will now analyze the properties of \ourshort{}s representing Lambertian 3D scenes, and illustrate how the geometry of the underlying 3D scene is encoded.
We will first derive an expression that establishes a relationship between \ourshort{}s and the classic two-plane parameterization of the light field.
Subsequently, we will derive an expression for the depth of a ray in terms of the local color gradient of the light field, therefore allowing us to efficiently extract sparse depth maps from the light field at any camera pose via analytical differentiation of the neural implicit representation. Please see \autoref{fig:2D_light_field} for an overview.

\paragraph{Locally linear slices of the light field.}
We derive here a local parametrization that will allow us to work with an \ourshort{} as if it were a conventional $2$-plane light field.
Given a ray $\ray$ in \plucker{} coordinates, we pick two points  $\mathbf{x}, \mathbf{x'} \in \R^3$ along this ray.
We then find a normalized direction $\dir \in \mathbb{S}^2$ not parallel to the ray direction - a canonical choice is a direction orthogonal to the ray direction.
We may now parameterize two parallel lines $\linea(s)= \mathbf{x} + s \dir$ and $\lineb(t) = \mathbf{x'} + t\dir$ that give rise to a local two-plane basis of the light field with ray coordinates $s$ and $t$.
$\ray$ intersects these lines at the two-plane coordinates $(s,t)=(0,0)$. This choice of local basis now assigns the two-plane coordinates $(s,t)$ to the ray $\ray$ from $\linea(s)$ to $\lineb(t)$.
In \autoref{fig:2D_light_field}, we illustrate this process on a simple 2D scene.

\paragraph{Epipolar Plane Images and their geometry.}
The \plucker{} coordinates (see Eq. \ref{eq:plucker}) enable us to extract a 2D slice from an \ourshort{}  field by varying $(s,t)$ and sampling $\LF$ on the Pl\"ucker coordinates of the rays parametrized pairs of points on the lines $\linea(s)$ and $\lineb(t)$:
\begin{equation}
    \label{eq:two-plane-to-plucker}
    \ourc(s,t) = \LF\left(\mathbf{r}(s,t)\right), \text{where}\; \mathbf{r}(s,t) = \overrightarrow{\mathbf{a}(s)\mathbf{b}(t)} = \left( \frac{\mathbf{b}(t)-\mathbf{a}(s)}{\|\mathbf{b}(t)-\mathbf{a}(s)\|}, \frac{\mathbf{a}(s) \times \mathbf{b}(t)}{\|\mathbf{b}(t)-\mathbf{a}(s)\|}\right).
\end{equation} 
The image of this 2D slice $\mathbf{c}(s,t)$ is well-known in the light field literature as an \emph{Epipolar Plane Image} (EPI)~\cite{bolles1987epipolar}. 
EPIs carry rich information about the geometry of the underlying 3D scene.
For example, consider a point $\point$ on the surface of an object in the scene; please see \autoref{fig:2D_light_field} for a diagram.
A point $\point \in \R^2$  has a $1$-dimensional family of rays going through the point, which correspond to a (green) line $L_\point$ in the EPI.
In a Lambertian scene, all rays that meet in this point and that are not occluded by other objects must observe the same color.
Therefore, the light field is constant along this line. As one travels along $L_\point$, rotating through the family of rays through $\point$, one eventually reaches a (magenta) \emph{tangent ray} $\tau$  to the object.
At a tangent ray, the value of the EPI ceases to be constant, and the light field changes its color to whatever is disoccluded by the object at this tangent ray.
Because objects of different depth undergo differing amounts of parallax,
\begin{wrapfigure}{r}{0.5\textwidth}
	\vspace{-0.1in}
	\centering
	\includegraphics[width=0.5\textwidth]{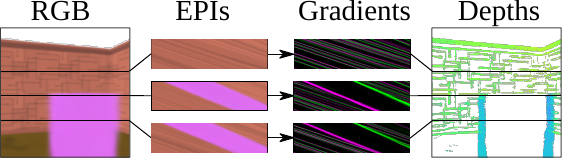}
	\label{fig:depth_estimation}
	\vspace{-0.3in}
\end{wrapfigure}

the \emph{slope} of the segment of $L_\point$ along which $\mathbf{c}$ is constant determines the 3D coordinates of $\point$. 
Finally, by observing that we may extract EPIs from \emph{any} perspective, it is clear that an \ourshort{}
encodes the full 3D geometry of the underlying scene.
Intuitively, this may also be seen by considering that one could render out all possible perspectives of the underlying scene, and solve a classic multi-view stereo problem to retrieve the shape.

\paragraph{Extracting depth maps from \ourshort{}s.}
A correctly inferred light field necessarily contains accurate 3D geometry information, although the geometry is encoded in a nontrivial way. To extract 3D geometry from an \ourshort{}, we utilize the property of the $2$-plane parameterization that the light field is constant on segments $L_\point$, the slopes of which determine $\point$. 
In the supplemental material, we derive 

\begin{proposition}
For a Lambertian scene, the distance $d$ along $\textbf{r} = \overrightarrow{\linea(s)\lineb(t)}$ from $\linea(s)$ to the point $\point$ on the object is 
\begin{equation}
    \label{eq:depth-along-ray}
    d(\mathbf{r}) = D \frac{ \partial_t \ourc(s,t)}{\partial_s \ourc(s,t) + \partial_t \ourc(s,t)}.
\end{equation}
where $\mathbf{a}(s)$ and $\mathbf{b}(t)$ are as above, $\ourc(s,t)$ is defined by (\ref{eq:two-plane-to-plucker}), $D$ is the distance between the lines $\mathbf{a}(t)$ and $\mathbf{b}(t)$. Thus  $\point = \linea(s) + d(\mathbf{r}) \frac{\lineb(t)-\linea(s)}{\|\lineb(t)-\linea(s)\|}$, and $\partial_x$ denotes the partial derivative by variable $x$.
\end{proposition}
This result yields meaningful depth estimates wherever the derivatives of the light fields are nonzero along the ray.
In practice, we sample several rays in a small $(s,t)$ neighborhood of the ray $\ray$ and declare depth estimates as invalid if the gradients have high variance-please see the code for implementation details.
This occurs when $\ray$ hits the object at a point where the surface color is changing, or when $\ray$ is a tangent ray.
We note that there is a wealth of prior art that could be used to extend this approach to extract \emph{dense} depth maps~\cite{dansereau2004gradient,tosic2014light, kim2013scene, si2016dense}.

\hypertarget{sec:metalearning}{}
\subsection{Meta-learning with conditional \ours{}s}
\label{sec:metalearning}
We consider a dataset $\dataset$ consisting of $N$ 3D scenes
\begin{equation}
    S_i = \{(\image_j, \extrinsiccamera_j, \intrinsiccamera_j)\}_{j=1}^K \in \R^{H \times W \times 3} \times SE(3) \times \R^{3 \times 3}, \; i=1\ldots N
\end{equation} 
with $K$ images $\image_j$ of each scene taken with cameras with extrinsic parameters $\extrinsiccamera_j$ and intrinsic parameters $\intrinsiccamera_j$~\cite{Hartley:2003}. 
Each scene is completely described by the parameters $\lfparameter_i \in \R^\ell$ of its corresponding light field MLP $\LF_i = \LF_{\lfparameter_i}$.

\paragraph{Meta-learning and multi-view consistency.}
In the case of 3D-structured neural scene representations, ray-marching or volumetric rendering naturally ensure multi-view consistency of the reconstructed 3D scene representation.
In contrast, a general 4D function $\Phi: \mathcal{L} \to \R^3$ is not multi-view consistent, as most such functions are not the light fields of any 3D scene.
We propose to overcome this challenge by learning a prior over the space of light fields. As we will demonstrate, this prior can also be used to reconstruct an \ourshort{} from a single 2D image observation.
In this paradigm, differentiable ray-casting is a method to force the light field of a scene to be multi-view consistent, while \emph{we instead impose multi-view consistency by learning a prior over light fields}.

\paragraph{Meta-learning framework.}
We propose to represent each 3D scene $S_i$ by its own latent vector 
 $\latent_i \in \R^{k}$. 
Generalizing to new scenes amounts to learning a prior over the space of light fields that is concentrated on the manifold of multi-view consistent light fields of natural scenes. 
To represent this latent manifold, we utilize a hypernetwork~\cite{ha2016hypernetworks,sitzmann2019srns}. The hypernetwork is a function, represented as an MLP
\begin{equation}
   \hypernetwork: \R^k \to \R^\ell, \hypernetwork_\psi(\latent_i) = \lfparameter_i
\end{equation}
with parameters $\psi$ which sends the latent code $\mathbf{z}_i$ of the $i$-th scene to the parameters of the corresponding \ourshort.

Several reasonable approaches exist to obtain latent codes $\latent_i$. One may leverage a convolutional- or transformer-based image encoder, directly inferring the latent from an image~\cite{kosiorek2021nerf,Niemeyer2020DVR}, or utilize gradient-based meta-learning~\cite{sitzmann2020metasdf}.
Here, we follow an auto-decoder framework~\cite{park2019deepsdf,sitzmann2019srns} to find the latent codes $\latent_i$, but note that \ourshort{}s are in no way constrained to this approach. We do not claim that this particular meta-learning method will out-perform other forms of conditioning, such as gradient-based meta-learning \cite{finn2017gradientbasedml, sitzmann2020metasdf} or FILM conditioning \cite{perez2018film}, but perform a comparison to a conditioning-by-concatenation approach in the appendix.
We assume that the latent vectors have a Gaussian prior with zero mean and a diagonal covariance matrix.  At training time, we jointly optimize the latent parameters $\latent_i$ together with the hypernetwork parameters $\hypernetworkparameter$ using the objective 
\begin{equation}
    \label{eq:training-objective}
 \argmin_{\{\latent_i\}, \hypernetworkparameter} \sum_i \sum_j \|\renderingmap^{\LF}_{\extrinsiccamera_j, \intrinsiccamera_j}(\Psi_{\hypernetworkparameter}(\latent_i))-\image_j\|_2^2 + \lambda_{lat}\|\latent_i\|^2_2.
 \end{equation}
Here the $\renderingmap^{\LF}$ is the rendering function (\autoref{eq:renderer}), the first term is an $\ell_2$ loss penalizing the light fields that disagree with the observed images, and the second term enforces the prior over the latent variables. We solve \autoref{eq:training-objective} using gradient descent. 
At test time, we freeze the parameters of the hypernetwork and reconstruct the light field for a new scene $S$ given a \emph{single observation} of the scene
$\{(\image, \extrinsiccamera, \intrinsiccamera)\}$ by optimizing, using gradient descent, the latent variable $\latent_S$ of the scene, such that the reconstructed light field $\LF_{\hypernetwork_\hypernetworkparameter(\latent_S)}$ best matches the given observation of the scene: 
\begin{equation}
    \latent_S = \argmin_{\latent}  \|\renderingmap^\LF_{\extrinsiccamera, \intrinsiccamera}\left(\Psi_\hypernetworkparameter(\latent) \right)-\image)\|_2^2 + \lambda_{lat}\|\latent\|^2_2.
\end{equation} 

\paragraph{Global vs. local conditioning}
The proposed meta-learning framework globally conditions an \ourshort{} on a single latent variable $\mathbf{z}$.
Recent work instead leverages \emph{local} conditioning, where a neural field is conditioned on local features extracted from a context image~\cite{saito2019pifu,yu2020pixelnerf,trevithick2020grf}.
In particular, the recently proposed pixelNeRF~\cite{yu2020pixelnerf} has achieved impressive results on few-shot novel view synthesis. As we will see, the current formulation of \ourshort{}s does \emph{not} outperform pixelNeRF.
We note, however, that local conditioning methods solve a \emph{different problem}. Rather than learning a prior over classes of \emph{objects}, local conditioning methods learn priors over \emph{patches}, answering the question ``How does this image patch look like from a different perspective?''.
As a result, this approach does not learn a latent space of neural scene representations. Rather, scene context is required to be available at test time to reason about the underlying 3D scene, and the representation is not \emph{compact}: the size of the conditioning grows with the number of context observations. In contrast, globally conditioned methods~\cite{sitzmann2019srns,kosiorek2021nerf,park2019deepsdf,mescheder2019occupancy} first infer a \emph{global} representation that is invariant to the number of context views and subsequently discard the observations.
However, local conditioning enables better generalization due to the shift-equivariance of convolutional neural networks.
An equivalent to local conditioning in light fields is non-obvious, and an exciting direction for future work.

\begin{figure*}
	\centering
	\includegraphics[width=\textwidth]{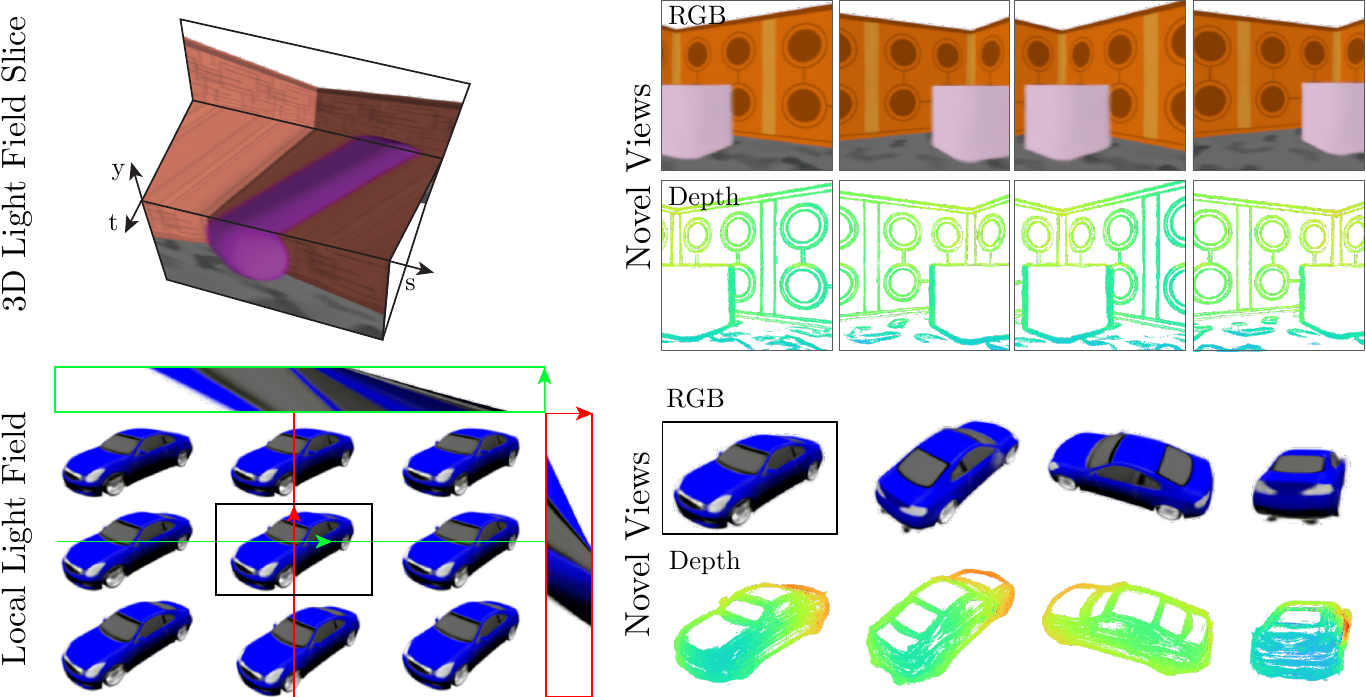}
	\caption{\textbf{360-degree light field parameterization.}
	Top left: 3D slice in the style of the Lumigraph~\cite{gortler1996lumigraph} of an \ourshort{} of a room-scale scene.
	Bottom left: nearby views of a car as well as vertical (right, red) and horizontal (top, green) Epipolar Plane Images, sampled from an \ourshort{}.
	Reconstructed from training set, 50 and 15 views, respectively.
	Right: \ourshort{}s enable rendering from arbitrary, 360-degree camera perspectives as well as sparse depth map extraction from only a single sample per ray.
	Please see the supplemental video for more qualitative results.
	}
	\label{fig:geometry}
	\vspace{-13pt}
\end{figure*}

\section{Experiments}
We demonstrate the efficacy of \ourshort{}s by reconstructing 360-degree light fields of a variety of simple 3D scenes.
In all experiments, we parameterize \ourshort{}s via a 6-layer ReLU MLP, and the hypernetwork as a 3-layer ReLU MLP, both with layer normalization. We solve all optimization problems using the ADAM solver with a step size of $10^{-4}$.
Please find more results, as well as precise hyperparameter, implementation, and dataset details, in the supplemental document and video.

\paragraph{Reconstructing appearance and geometry of single-object and room-scale light fields.}
We demonstrate that \ourshort{} can parameterize 360-degree light fields of both single-object ShapeNet~\cite{chang2015shapenet} objects and simple, room-scale environments.
We train \ourshort{}s on the ShapeNet ``cars'' dataset with 50 observations per object from ~\cite{sitzmann2019srns}, as well as on simple room-scale environments as proposed in~\cite{eslami2018neural}.
Subsequently, we evaluate the ability of \ourshort{}s to generate novel views of the underlying 3D scenes. Please see \autoref{fig:geometry} for qualitative results.
\ourshort{}s succeed in parameterizing the 360-degree light field, enabling novel view synthesis at \emph{real-time} frame-rates (see supplemental video).
We further demonstrate that \ourshort{}s encode scene geometry by presenting Epipolar Plane Images and leveraging the relationship derived in \autoref{eq:depth-along-ray} to infer sparse depth maps.
We highlight that both rendering and depth map extraction \emph{do not require ray-casting}, with only a single evaluation of the network or the network and its gradient respectively.

\begin{figure*}[t!]
	\centering
	\includegraphics[width=\textwidth]{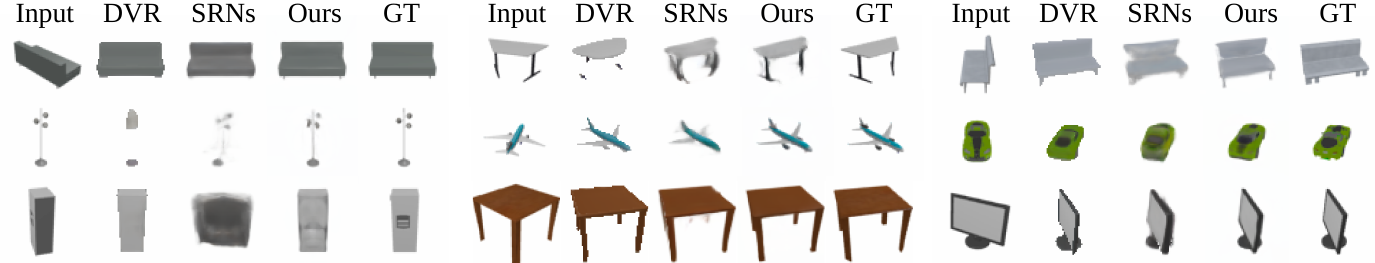}
	\vspace{-0.4cm}
	\caption{\textbf{Category-agnostic single-shot reconstruction.} We train a single model on the 13 largest Shapenet classes. \ourshort{}s consistently outperform global conditioning baselines. Baseline results courtesy of the pixelNeRF~\cite{yu2020pixelnerf} authors. Please see the supplemental video for more qualitative results.}
	\label{fig:nmr}
\end{figure*}
\begin{table}
\caption{\textbf{Single-shot multi-class reconstruction results.} We benchmark \ourshort{}s with SRNs~\cite{sitzmann2019srns} and DVR~\cite{Niemeyer2020DVR} on the task of single-shot (auto-decoding with a single view), multi-class reconstruction of the 13 largest ShapeNet~\cite{chang2015shapenet} classes. \ourshort{}s significantly outperform DVR and SRNs on almost all classes, on average by more than $1$dB, while only requiring a \emph{single} network evaluation per ray. Note that DVR requires additional supervision in the form of foreground-background masks. Means across classes are weighted equally.}
\vspace{0.1in}
\resizebox{\textwidth}{!}{
\begin{tabular}{@{}llllllllllllllll@{}}
\toprule
                   &      & plane          & bench          & cbnt.          & car            & chair          & disp.          & lamp           & spkr.          & rifle          & sofa           & table          & phone          & boat           & mean           \\ \midrule
\multirow{3}{*}{$\uparrow$ PSNR}    & DVR~\cite{Niemeyer2020DVR}  & 25.29          & 22.64          & 24.47          & 23.95          & 19.91          & \bf{20.86}          & 23.27          & 20.78          & 23.44          & 23.35          & 21.53          & \bf{24.18}          & 25.09          & 22.70          \\
                   & SRN~\cite{sitzmann2019srns}  & 26.62          & 22.20          & 23.42          & 24.40          & 21.85          & 19.07          & 22.17          & 21.04          & 24.95          & 23.65          & 22.45          & 20.87          & 25.86          & 23.28          \\
                & \ourshort{} & \bf{29.95} & \bf{23.21} & \bf{25.91} & \bf{28.04} & \bf{22.94} & 20.64 & \bf{24.56} & \bf{22.54} & \bf{27.50} & \bf{25.15} & \bf{24.58} & 22.21 & \bf{27.16} & \bf{24.95} \\ \midrule
 \multirow{3}{*}{$\uparrow$ SSIM}  & DVR~\cite{Niemeyer2020DVR}  & 0.905          & \bf{0.866}          & \bf{0.877}          & 0.909          & 0.787          & \bf{0.814}          & \bf{0.849}          & 0.798          & 0.916          & 0.868          & 0.840          & \bf{0.892}          & 0.902          & 0.860          \\
                   & SRN~\cite{sitzmann2019srns}  & 0.901          & 0.837          & 0.831          & 0.897          & 0.814          & 0.744          & 0.801          & 0.779          & 0.913          & 0.851          & 0.828          & 0.811          & 0.898          & 0.849          \\
                   & \ourshort{} & \bf{0.932} & 0.855 & 0.871 & \bf{0.943} & \bf{0.835} & 0.786 & 0.844 & \bf{0.808} & \bf{0.939} & \bf{0.874} & \bf{0.868} & 0.844 & \bf{0.907} & \bf{0.870} \\
\midrule
\end{tabular}
}
\label{tbl:nmr}
\vspace{-10pt}
\end{table}
\paragraph{Multi-class single-view reconstruction.}
Following~\cite{Niemeyer2020DVR, yu2020pixelnerf}, we benchmark \ourshort{}s with recent \emph{global conditioning methods} on the task of single-view reconstruction and novel view synthesis of the 13 largest ShapeNet categories. We follow the same evaluation protocol as~\cite{kato2018neural} and train a single model across all categories.
See \autoref{fig:nmr} for qualitative and \autoref{tbl:nmr} for quantitative baseline comparisons. We significantly outperform both Differentiable Volumetric Rendering (DVR)~\cite{Niemeyer2020DVR} and Scene Representation Networks (SRNs)~\cite{sitzmann2019srns} on all but two classes by an average of 1dB, while requiring more than an order of magnitude fewer network evaluations per ray. Qualitatively, we find that the reconstructions from \ourshort{}s are often crisper than those of either Scene Representation Networks or DVR. Note that DVR requires additional ground-truth foreground-background segmentation masks.

\begin{figure*}
	\begin{minipage}[t]{0.53\textwidth}
    	\vspace{0pt}
    	\centering
    	\includegraphics[width=\textwidth]{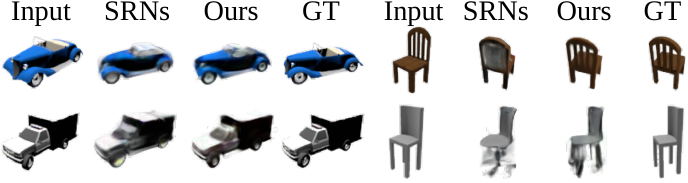}
	\end{minipage} \hfill           
	\begin{minipage}[t]{0.47\textwidth}
    	\vspace{10pt}
	    \centering
	    \small
    	\begin{tabular}{lcc}
    		\toprule
    		& Chairs 	& Cars \\ 
    		\midrule
    		SRNs~\cite{sitzmann2019srns}
    		& $\mathbf{22.89}$ / $0.89$ & $22.25$ / $\mathbf{0.89}$ \\
    		\ourshort{}
    		& $22.26$ / $\mathbf{0.90}$	& $\mathbf{22.42}$ / $\mathbf{0.89}$ \\
    		\bottomrule
    	\end{tabular}
	\end{minipage}
	\caption{\textbf{Class-specific single-shot reconstruction.} \ourshort{} performs approximately on par with SRNs~\cite{sitzmann2019srns} in the single-class single-shot reconstruction case, while requiring an order of magnitude fewer network evaluations, memory, and rendering time. Quantitative results report (PSNR, SSIM).}
	\label{fig:srns}
\end{figure*}

\paragraph{Class-specific single-view reconstruction.}
We benchmark~\ourshort{}s on single-shot reconstruction on the Shapenet ``cars'' and ``chairs'' classes as proposed in SRNs~\cite{sitzmann2019srns}.
See \autoref{fig:srns} for qualitative and quantitative results. We report performance better than SRNs in PSRN and on par in terms of SSIM on the ``cars'' class, and  worse in PSNR but better in terms of SSIM on the ``chairs'' class, while requiring an order of magnitude fewer network evaluations and rendering in real-time. We attribute the drop in performance compared to multi-class reconstruction to the smaller dataset size, causing multi-view inconsistency.

\begin{figure*}
	\begin{minipage}[t]{0.55\textwidth}
    	\vspace{0pt}
    	\centering
    	\includegraphics[width=\textwidth]{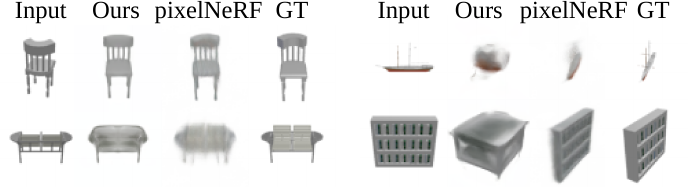}
	\end{minipage} \hfill           
	\begin{minipage}[t]{0.45\textwidth}
    	\vspace{10pt}
	    \centering
	    \small
    	\begin{tabular}{lcc}
    		\toprule
    		& Class-specific & Multi-Class \\ 
    		\midrule
    		\ourshort{}
    		& $22.34$ / $0.90$	& $24.95$ / $0.87$ \\
    		pixelNeRF
    		& $\mathbf{23.45}$ / $\mathbf{0.91}$ & $\mathbf{26.80}$ / $\mathbf{0.91}$ \\
    		\bottomrule
    	\end{tabular}
	\end{minipage}
	\caption{\textbf{Global vs. local conditioning.} The locally conditioned pixelNeRF outperforms \ourshort{}s in single-shot reconstruction, though \ourshort{}s require \emph{three orders of magnitude less rendering time and memory}. Qualitatively, for many objects \ourshort{}s are on par with pixelNERF (left), but confuse the object class on others (right). Quantitative results report (PSNR, SSIM). `Class-specific' refers to average over cars and chairs as in Fig. \ref{fig:srns} while 'multi-class' refers to the average over all ShapeNet classes as in Fig. \ref{tbl:nmr}.}
	\label{fig:pixelnerf}
\end{figure*}

\paragraph{Global vs. local conditioning and comparison to pixelNeRF~\cite{yu2020pixelnerf}.}
We investigate the role of global conditioning, where a single latent is inferred to describe the whole scene~\cite{sitzmann2019srns}, to local conditioning, where latents are inferred \emph{per-pixel} in a 2D image and leveraged to locally condition a neural implicit representation~\cite{saito2019pifu,trevithick2020grf,yu2020pixelnerf}.
We benchmark with the recently proposed pixelNeRF~\cite{yu2020pixelnerf}.
As noted above (see \hyperlink{sec:metalearning}{Section \ref{sec:metalearning}}), local conditioning does \emph{not} infer a compact neural scene representation of the  scene.
Nevertheless, we provide the comparison here for completeness.
See \autoref{fig:pixelnerf} for qualitative and quantitative results. 
On average, \ourshort{}s perform 1dB worse than pixelNeRF in the single-class case, and 2dB worse in the multi-class setting. %

\begin{table*}
	\centering
	\caption{\textbf{Comparison of rendering complexity.} \ourshort{}s require \emph{three orders of magnitude} less compute than volumetric  rendering based approaches, and admit real-time rendering. Please see Table  of supplemental material for comparison on larger images based on data in \cite{}}
	\small
	\begin{tabular}{lcccc}
		\toprule
		& \ourshort{}s & SRNs~\cite{sitzmann2019srns} &   pixelNeRF~\cite{yu2020pixelnerf} \\ 
		\midrule
		\# evaluations per ray
		& 1 & 11 & 192 \\
		clock time for $256\times256$ image (ms) 
		& 2.1 & 120 & $30e3$\\
		\bottomrule
	\end{tabular}
	\label{tbl:rendering_complexity}
\end{table*}

\paragraph{Real-time rendering and storage cost.}
See \autoref{tbl:rendering_complexity} for a quantitative comparison of the rendering complexity of \ourshort{} compared with that of volumetric and ray-marching based neural renderers~\cite{sitzmann2019srns, yariv2020multiview,lombardi2019neural,mildenhall2020nerf,yu2020pixelnerf}.  %
All clock times were collected for rendering $256\times256$ images on an NVIDIA RTX 6000 GPU.
We further compare the cost of storing a single \ourshort{} with the cost of storing a conventional light field.
With approximately $400$k parameters, a single \ourshort{} requires around $1.6$ MB of storage, compared to $146$ MB required for storing a 360-degree light field at a resolution of $256\times256\times17\times17$ in the six-plane Lumigraph configuration.

\paragraph{Multi-view consistency as a function of training set size.}
We investigate how multi-view consistency scales with the amount of data that the prior is trained on. Please find this analysis in the supplementary material.

\paragraph{Overfitting of single 3D scenes.}
We investigate overfitting a single 3D scene with a Light Field Network with positional encodings / sinusoidal activations~\cite{sitzmann2019siren,tancik2020fourier}. Please find this analysis in the supplementary material.

\begin{figure*}
	\begin{minipage}[t]{0.71\textwidth}
    	\vspace{0pt}
    	\centering
    	\includegraphics[width=\textwidth]{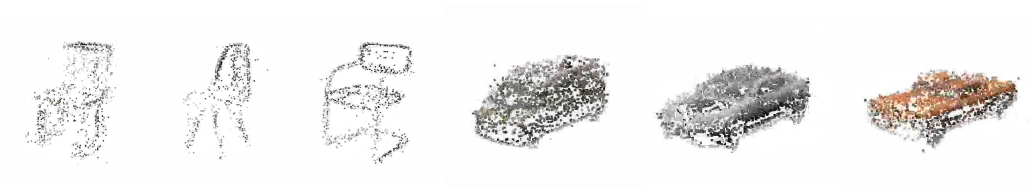}
	\end{minipage} \hfill           
	\begin{minipage}[t]{0.28\textwidth}
    	\vspace{10pt}
	    \centering
	    \small
        \begin{tabular}{lcc}
            \toprule
            Method & Cars & Chairs \\
        	\midrule
            SRNs 	& 0.071 & 0.20 \\
            \ourshort{} 	& \textbf{0.059} & \textbf{0.130} \\
        	\bottomrule
        \end{tabular}
        \label{tbl:depth}
	\end{minipage}
    \caption{\textbf{Geometry reconstruction with LFNs.} By backprojecting extracted depth into 3D, LFNs enable reconstruction of 3D pointclouds (left). For rays were depth is valid, \ourshort{}s achieve lower mean L1 depth error than Scene Representation Networks (right).}
	\label{fig:depth}
\end{figure*}%
\paragraph{Evaluation of Reconstructed Geometry.}
We investigate the quality of the geometry that can be computed from an \ourshort{} via Eq.~\ref{eq:depth-along-ray}. For every sample in the class-specific single-shot reconstruction experiment, we extract its per-view sparse depth map. We then backproject depth maps from four views into 3D to reconstruct a point cloud, and benchmark mean L1 error on valid depth estimates with Scene Representation Networks~\cite{sitzmann2019srns}. Fig.~\ref{fig:depth} displays qualitative and quantitative results. Qualitatively, point clouds succeed in capturing fine detail such as the armrests of chairs. Quantitatively, LFNs outperform SRNs on both cars and chairs. We note that LFNs have a slight advantage in this comparison, as we can only benchmark on the sparse depth values, for which LFNs have high confidence. This includes occlusion boundaries, which are areas where the sphere-tracing based SRNs incurs high error, as it is forced to take smaller and smaller steps and may not reach the surface. We highlight that we do not claim that the proposed method is competetive with methods designed for geometry reconstruction in particular, but that we only report this to demonstrate that the proposed method is capable to extract valid depth estimates from an LFN.

\paragraph{Limitations.}
\begin{wrapfigure}{r}{0.4\textwidth}
\label{fig:occluder}
	\vspace{-0.4in}
	\centering
    \includegraphics[width=0.4\textwidth]{"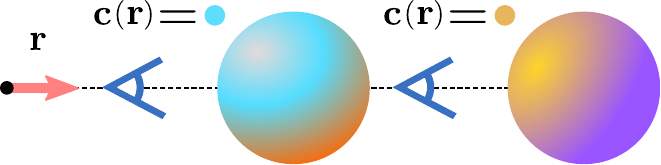"}
	\vspace{-0.25in}
\end{wrapfigure}

First, as every existing light field approach, \ourshort{}s store only one color per oriented ray, which makes rendering views from cameras placed in between occluding objects challenging, even if the information may still be stored in the light field.
Second, though we outperform globally-conditioned methods, we currently do not outperform the locally conditioned pixelNeRF.
Finally, as opposed to 3D-structured representations, \ourshort{}s do not enforce strict multi-view consistency, and may be inconsistent in the case of small datasets.

\section{Discussion and Conclusion}
We have proposed \ours{}s, a novel neural scene representation that directly parameterizes the full 360-degree, 4D light field of a 3D scene via a neural implicit representation.
This enables both real-time neural rendering with a \emph{single} evaluation of the neural scene representation per ray, as well as sparse depth map extraction without ray-casting.
\ours{}s outperform globally conditioned baselines in single-shot novel view synthesis, while being three orders of magnitude faster and less memory-intensive than current volumetric rendering approaches.
Exciting avenues for future work include combining \ourshort{}s with local conditioning, which would enable stronger out-of-distribution generalization, studying the learning of non-Lambertian scenes, and enabling camera placement in obstructed 3D space.
With this work, we make important contributions to the emerging fields of neural rendering and neural scene representations, with exciting applications across computer vision, computer graphics, and robotics.

\paragraph{Societal Impacts.}
Potential improvements extending our work on few-observation novel view synthesis could enable abuse by decreasing the cost of non-consensual impersonations. We refer the reader to a recent review of neural rendering \cite{tewari2020state} for an in-depth discussion of this topic.

\section*{Acknowledgements and Disclosure of Funding}
This work is supported by the NSF under Cooperative Agreement PHY-2019786 (The NSF AI Institute for Artificial Intelligence and Fundamental Interactions, \url{http://iaifi.org/}), 
ONR under 1015 G TA243/N00014-16-1-2007 (Understanding Scenes and Events through Joint Parsing, Cognitive Reasoning and Lifelong Learning), Mitsubishi under 026455-00001 (Building World Models from some data through analysis by synthesis), DARPA under CW3031624 (Transfer, Augmentation and Automatic Learning with Less Labels), as well as the Singapore DSTA under DST00OECI20300823 (New Representations for Vision).
We thank Andrea Tagliasacchi, Tomasz Malisiewicz, Prafull Sharma, Ludwig Schubert, Kevin Smith, Bernhard Egger, Christian Richardt, Manuel Rey Area, and J\"{u}rgen and Susanne Sitzmann for interesting discussions and feedback, and Alex Yu for kindly sharing the outputs of pixelNeRF and baselines with us.

{\small
	\bibliographystyle{unsrt}
	\bibliography{references,macros}
}



\section*{Supplementary material}

# Light Field Networks: Neural Scene Representations with Single-Evaluation Rendering
# –Supplementary Material–

**Vincent Sitzmann**[1,*]
sitzmann@mit.edu

**Semon Rezchikov**[2,*]
skr@math.columbia.edu

**William T. Freeman**[1,3]
billf@mit.edu

**Joshua B. Tenenbaum**[1,4,5]
jbt@mit.edu

**Frédo Durand**[1]
fredo@mit.edu

[1]MIT CSAIL  [2]Columbia University  [3] IAFI  [4]MIT BCS  [5] CBMM

vsitzmann.github.io/lfns/

## Contents

---

[*]These authors contributed to this work.

# 1 Additional results on the local two-plane parameterization

In this section, we derive Proposition 1 of the paper. We recall here that the Plücker coordinates parameterize the space $\mathcal{L}$ of oriented lines in $\mathbb{R}^3$ as the set of tuples

$$\mathcal{L} = \{(\mathbf{d}, \mathbf{w}) \mid \mathbf{d}, \mathbf{w} \in \mathbb{R}^3;\ \mathbf{d} \cdot \mathbf{w} = 0, \|\mathbf{d}\| = 1\}. \tag{1}$$

The line through $\mathbf{x}$ in direction $\mathbf{d}$ (with $\|\mathbf{d}\| = 1$), i.e. the line $\overline{\mathbf{x}, \mathbf{x} + \mathbf{d}}$, has normalized Plücker coordinates

$$(\mathbf{d}, \mathbf{x} \times (\mathbf{x} + \mathbf{d})) = (\mathbf{d}, \mathbf{x} \times \mathbf{d}).$$

If the direction vector $\mathbf{d}$ is not normalized, we can still compute Plücker coordinates as above; to normalize the coordinates we must apply the normalization operator

$$N(\mathbf{d}, \mathbf{w}) = (\mathbf{d}, \mathbf{w})/\|\mathbf{d}\|. \tag{2}$$

Thus the ray $\overrightarrow{\mathbf{a}(s)\mathbf{b}(t)}$ has (unnnormalized) Plücker coordinates

$$v(s, t) = (\mathbf{b} - \mathbf{a}, \mathbf{a} \times \mathbf{b}). \tag{3}$$

*Proof of Proposition 1.* Given two coplanar lines

$$\mathbf{a}(s) = \mathbf{x} + s\mathbf{d} \text{ and } \mathbf{b}(t) = \mathbf{x}' + t\mathbf{d} \tag{4}$$

we write

$$\mathbf{c}(s, t) = \Phi\left(\frac{\mathbf{b} - \mathbf{a}}{\|\mathbf{b} - \mathbf{a}\|}, \frac{\mathbf{a} \times \mathbf{b}}{\|\mathbf{b} - \mathbf{a}\|}\right) = \Phi(N(v(s, t))) \tag{5}$$

for the color of the ray $\ell = \overrightarrow{\mathbf{a}(s)\mathbf{b}(t)}$ described by the LFN $\Phi$. Suppose $\ell$ captures the color of a point $\mathbf{p}$ in a 3D scene. For a Lambertian scene, if the ray is not a tangent ray, $\mathbf{c}(s, t)$ should be constant near $\ell$ along the line in the $(s, t)$ plane corresponding to the family of rays through $\mathbf{p}$. *Therefore, the gradient $\nabla_{s,t}\mathbf{c}(s, t)$ is orthogonal to this line in the $(s, t)$ plane.*

Write $J$ for the matrix of rotation by 90 degrees, i.e. $[0, -1; 1, 0]$. Thus at non-tangent lines, $J\nabla_{s,t}\mathbf{c}(s, t)$ will point along the family of rays through $\mathbf{p}$.

We write $D$ for the distance between the lines $\mathbf{a}$ and $\mathbf{b}$; so $D = \|\mathbf{x}' - \mathbf{x}\|$ for $\mathbf{a}, \mathbf{b}$ as in (4). Let $\mathbf{a}_0, \mathbf{b}_0$ be the closest points to $\mathbf{p}$ on $\mathbf{a}(s)$ and $\mathbf{b}(t)$, respectively. Write $\mathbf{a}_1 = \mathbf{a}(s)$ and $\mathbf{b}_1 = \mathbf{b}(t)$ for the intersection of $\ell$ with $\mathbf{a}, \mathbf{b}$; then a nearby ray on the pencil of rays through $\mathbf{p}$ is given by $\ell' = \overrightarrow{\mathbf{a}'_1\mathbf{b}'_1}$ for some $\mathbf{a}'_1 = \mathbf{a}(s - \delta s), \mathbf{b}'_1 = \mathbf{b}(t + \delta t)$. We have similar triangles $\triangle \mathbf{p}\mathbf{a}_0\mathbf{a}_1, \triangle \mathbf{p}\mathbf{a}_0\mathbf{a}'_1, \triangle \mathbf{p}\mathbf{b}_0\mathbf{b}_1$, and $\triangle \mathbf{p}\mathbf{b}_0\mathbf{b}'_1$. Since the $(s, t)$ coordinates of $\ell'$ differ from the $(s, t)$ coordinates of $\ell$ by a multiple of $J\nabla_{s,t}\mathbf{c}(s, t)$, the similarity relationships between the triangles above imply that

$$\frac{\delta s}{\delta t} = \frac{-(J\nabla \mathbf{c})_s}{(J\nabla \mathbf{c})_t} = \frac{d}{D - d}.$$

This simplifies to

$$\frac{\partial_t \mathbf{c}}{\partial_s \mathbf{c}} = \frac{d}{D - d}.$$

Rearranging, we have

$$d = D\frac{\partial_t \mathbf{c}}{\partial_s \mathbf{c} + \partial_t \mathbf{c}}. \tag{6}$$

□

Let $\delta$ be the distance from $\mathbf{a}(s)$ to $p$; then $\delta = \sqrt{s^2 + d^2}$.

Let's now describe how to get an estimate of a point generating the color of a ray in Plücker coordinates. Say we have (normalized) Plücker coordinates $(\mathbf{d}, \mathbf{w})$. Write $\mathbf{x} = \mathbf{d} \times \mathbf{w}$; this is the closest point on the line $\ell_{(\mathbf{d}, \mathbf{w})}$ to the origin. Now choose auxiliary $\mathbf{d}'$ with $\|\mathbf{d}\| = 1$. (Ideally $\mathbf{d}$ is

not close to $\mathbf{d}'$.) Write $\mathbf{x}' = \mathbf{x} + D\mathbf{d}$ for some positive number $D$. Then we consider the two-plane parametrization through

$$\mathbf{a}(s) = \mathbf{x} + s\mathbf{d}', \mathbf{b}(t) = \mathbf{x}' + t\mathbf{d}' \qquad (7)$$

Deine $\mathbf{c}$ via (5) where we define $v$ as in (3) using $\mathbf{a}, \mathbf{b}$ as in (7). The line $\overrightarrow{\mathbf{x}, \mathbf{x} + \mathbf{d}} = \ell_{(\mathbf{d}, \mathbf{w})}$ intersects $\mathbf{a}(s)$ at $\mathbf{x} = \mathbf{a}(0)$ and intersects $\mathbf{b}(t)$ at $\mathbf{x}' = \mathbf{x} + D\mathbf{d} = \mathbf{b}(0)$. Compute $\partial_s \mathbf{c}, \partial_t \mathbf{c}$ — a procedure that involves computing analytical derivatives of the neural network $\Phi$ — then compute $d(s,t)$ via (6), and then

$$\mathbf{p} = \mathbf{x} + \delta(0,0)\mathbf{d}.$$

## 2 Reproducibility

In the following, we provide all the details necessary to reproduce the experiments outlined in the paper. *All code and datasets will be made publicly available.*

### 2.1 Hardware

Each model was separately trained on a single NVIDIA RTX 6000 GPU with 24 GB of memory. Overall, we used up to 4 GPUs in parallel.

### 2.2 Architecture Details

All LFNs are implemented as six-layer fully connected neural networks with ReLU nonlinearities and 256 hidden units per layer. Before each layer, we leverage layer normalization *without affine transforms*, i.e., no additional parameters are introduced. All hypernetworks are implemented as three-layer fully connected neural networks with ReLU nonlinearities and layer normalization *with* affine transform. The hidden layer size of the hypernetworks is 256. Thus, the parameter $\ell$ in Equation 10 of the main text is $\sim 400,000$. The size of latent codes $\mathbf{z}$ is 256.

### 2.3 360-degree light field reconstruction experiments (Figure 5)

**Cars dataset.** We use the dataset proposed by Sitzmann et al. [9], hosted by the authors of pixelNeRF [11] under https://drive.google.com/drive/folders/1PsT3uKwqHHD2bEEHkIXB99AlIjtmrEiR.

**Rooms dataset.** Using the assets provided by the authors of GQN [2], hosted under https://github.com/deepmind/lab/tree/master/assets/textures/map/lab_games and a modified version of the Blender Shapenet rendering script hosted under https://github.com/vsitzmann/shapenet_renderer/blob/master/shapenet_spherical_renderer.py, we rendered $10,000$ rooms. The rooms have a sidelength of $7$. Cameras are placed exclusively in the center $2 \times 2$ square of the room. Objects are placed exclusively in the outer $1.5$ perimeter of the room, such that the *camera is only placed in unobstructed free space*. Colors and types of objects are chosen at random. Every room features between $1$ and $5$ objects. We render 30 observations per room.

**Experiment Details.** We train separate models for the "cars" and "rooms" classes. We train in two resolution stages. First, at a resolution of $64 \times 64$ and a batch size of 300, then, at a batch size of 75 at a resolution of $128 \times 128$ until convergence for a total of approx. 3 days, both using the ADAM optimizer with a step size of $10^{-4}$. We choose the parameter $\lambda_{lat}$ as $1e2$. At test time, we freeze the network parameters, initialize all latent codes to the prior mean (i.e., all zeros), and optimize the latent codes until convergence. Hyperparameters were discovered via unstructured search.

### 2.4 Multi-class single shot reconstruction

**Dataset.** We use the dataset proposed by Kato et al. [3], available for download under https://s3.eu-central-1.amazonaws.com/avg-projects/differentiable_

16

`volumetric_rendering/data/NMR_Dataset.zip` (hosted by the authors of Differentiable Volumetric Rendering [5]).

**Experiment Details.** We train a single model on all 13 classes. We train until convergence (approx. two days) at a resolution of $64 \times 64$ and a batch size of 300 until convergence for a total of approx. 3 days using the ADAM optimizer with a step size of $10^{-4}$. We choose the parameter $\lambda_{lat}$ as $1e2$. At test time, we freeze the network parameters, initialize all latent codes to the prior mean (i.e., all zeros), and optimize the latent codes until convergence. Leveraging an aggressive learning rate schedule, auto-decoding converges within 200 iterations or approximately 1 second for a batch of 300 shapes. Hyperparameters were discovered via unstructured search.

### 2.5 Single-class single shot reconstruction

**Dataset.** We use the dataset proposed by Sitzmann et al. [9], hosted by the authors of pixelNeRF [11] under https://drive.google.com/drive/folders/1PsT3uKwqHHD2bEEHkIXB99AlIjtmrEiR.

**Experiment Details.** We train separate models for the "cars" and "chairs" classes. We train in two resolution stages. First, at a resolution of $64 \times 64$ and a batch size of 300 for 200k steps, then, at a batch size of 75 at a resolution of $128 \times 128$ until convergence (total approx. 3 days), both using the ADAM optimizer with a step size of $10^{-4}$. We choose the parameter $\lambda_{lat}$ as $1e2$. At test time, we freeze the network parameters, initialize all latent codes to the prior mean (i.e., all zeros), and optimize the latent codes until convergence, as in the multi-class single-shot experiment. Hyperparameters were discovered via unstructured search.

## 3 Additional Results

In fig. 1 we show novel views of training set objects together with extracted epipolar plane images and depth maps. Please see the supplemental video for extensive qualitative results.

In Table 1 we show additional results comparing rendering complexity of LFNs to other existing methods, for larger, $800 \times 800$ pixel images.

**Comparison to conditioning-by-concatenation.** We performed an ablation study comparing hypernetworks to the simpler alternative of conditioning via concatenation. We parameterized the LFN identically to the MLPs in pixelNeRF and DVR: A 5-layer, fully-connected residual MLP with ReLU activations and 512 hidden units. The latent code was directly concatenated to the input ray coordinate. The remaining experimental settings were identical to those in Section 2.4 above. Both methods were trained for approximately 3 days on a single RTX 6000 GPU. In this setting, conditioning via concatenation yields significantly worse performance, as shown in Table 2. This is in line with the insight from Pi-GAN [1], where FILM-conditioning [6], which is an intermediate between a hypernetwork and conditioning via concatenation, was found to perform significantly better than conditioning via concatenation. Similarly, Facebook AI has found the hypernetwork-conditioned SRN to outperform concatenation-conditioned DVR and NeRF alternatives (see minute 16:30 of video associated to [7]). We further note that conditioning via concatenation is a special case of a hypernetwork, namely a single linear layer outputting the biases of the first LFN layer - see for instance appendix of [8] for a proof.

We note that we do not claim that the proposed method of conditioning is superior to any other method of conditioning. Rather, the proposed LFN framework is compatible with any conditioning method. We merely argue that to learn multi-view consistent light fields, we require some form of meta-learning, and see the analysis of the optimal conditioning method as an avenue for future work.

**Ablating overfitting vs. generalization.** To validate our claim that meta-learning on a dataset of 3D shapes enables consistent novel view synthesis on test objects by learning a multi-view consistency prior in light-fields, we benchmarked PSNR of novel views on 10 objects in the 10 set, reconstructed from a single observation using models trained on 1, 10, 100, 1k, and up to 2500 shapes, both in the single-class and multi-class settings. Experimental settings are as in Sections 2.4 and 2.5. Please

see Table 3 for results. The trend in PSNR is consistent improvement, suggesting that additional training-set objects would likely improve performance further.

We further note that meta-learning across a dataset of 3D scenes serves two distinct roles: learning a space of multi-view consistent light fields, and learning a prior over 3D scenes to enable few-shot reconstruction. To better disentangle these two roles, we ran a second experiment. For each instance in the training set, which was drawn from the ShapeNet chair class, we randomly picked 5 views, such that on average, almost every surface point of each chair has been observed, but not every oriented ray. I.e., while the 3D surfaces of the chairs are densely sampled, their light fields are not. We trained LFNs on 10, 100, 1000, and all (4.5k) chair instances, including the 10 chairs previously mentioned, and evaluated view synthesis performance in PSNR on a held-out set of novel views for each of these 10 chairs. This tests only the first property of the meta-learning approach - i.e., how the meta-learning learns multi-view consistency as an abstract property - without entangling it with the capability to perform few-shot reconstruction. Experimental parameters were as in 2.5. Please see Table 4 for results. The performance similarly improves logarithmically.

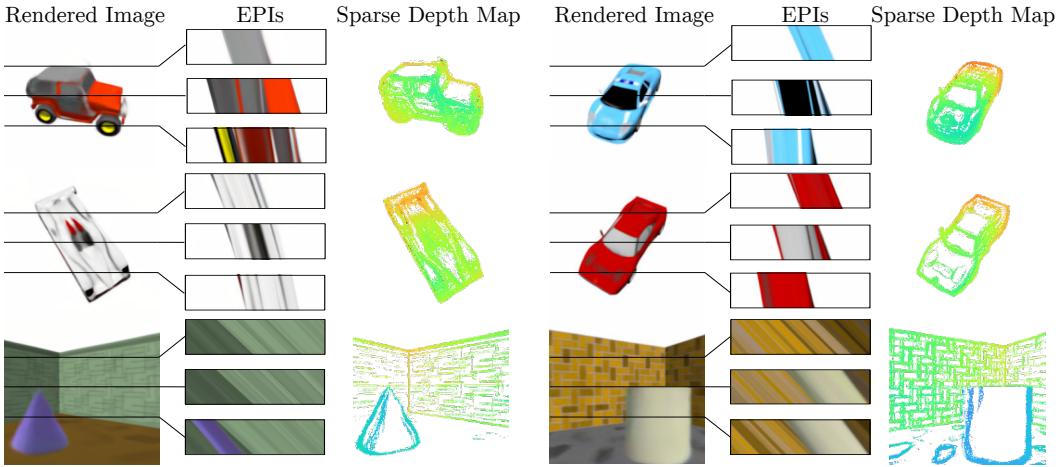

Figure 1: Novel views of cars and rooms with Epipolar Plane Images and depth maps.

Table 1: **Comparison of rendering complexity, continued.** Data for all architectures aside from LFN taken from [7]; see 17:00 of associated video.

|  | LFNs | SRNs [9] | DVR [5] | IDR [10] | NeRF [4] |
|---|---|---|---|---|---|
| clock time for $800 \times 800$ image (ms) | 20.5 | 1e3 | 9e4 | 1e5 | 2e4 |

Table 2: **Comparison of meta-learning approach.** We compared conditioning via a hypernetwork, used in the paper, to conditining-by-concatenation (CvC below) in the 13-class ShapeNet dataset. All results are PSNR in dB.

| Method | plane | bench | cbnt. | car | chair | disp. | lamp | spkr. | rifle | sofa | table | phone | boat | mean |
|---|---|---|---|---|---|---|---|---|---|---|---|---|---|---|
| Hypernetwork | 29.95 | 23.21 | 25.91 | 28.04 | 22.94 | 20.64 | 24.56 | 22.54 | 27.50 | 25.15 | 24.58 | 22.21 | 27.16 | 24.95 |
| CvC | 24.08 | 19.43 | 21.01 | 23.57 | 19.81 | 17.47 | 19.05 | 19.40 | 22.24 | 20.85 | 19.81 | 17.64 | 24.02 | 20.64 |

**Overfitting a densely sampled 3D scene.** We demonstrate that LFNs can in principle represent high-frequency information. We fit an LFN to all the views of the "Fern" scene from the NeRF [4] dataset, which consists of photos of a real-world scene captured with a DSLR camera. We add positional encodings to the plucker embedding, following NeRF. Please see qualitative results in Fig 2, as well as the supplemental video. LFNs succeed in reproducing all context images perfectly, proving that in principle, an LFN is capable of parameterizing realistic, 4d high-frequency content. As expected, rendering out the intermediate views leads to basically random images, please see this video, as there is no mechanism to enforce multi-view consistency, in contrast to the experiments in the paper, where this prior was learnt using meta-learning. We note that fundamentally, there are

Table 3: **Multi-view consistency vs. training set size.** As we scale the size of the training set, the multi-view consistency prior improves consistently, both in the single-class and multi-class setting. All results are PSNR in dB. Top row denotes number of object instances in the traninig set. The cars dataset contains approximately 2500 objects, while the NMR dataset contains approximately 2000 objects per class.

| Dataset | 1 | 10 | 100 | 1k | all |
|---|---|---|---|---|---|
| Cars (Single-class) | 16.21 | 18.98 | 20.76 | 23.09 | 23.56 |
| NMR (Multi-Class) | 18.03 | 19.77 | 20.82 | 23.87 | 24.95 |

Table 4: **Disentangling multi-view consistency vs. object prior.** Columns indicate number of instances of chair ShapeNet class. Each LFN was trained on 5 random views of each instance and evaluated on 10 novel views of a fixed subset of 10 training set chairs.

| | 10 | 100 | 1k | all |
|---|---|---|---|---|
| PSNR (dB) on novel views | 15.88 | 16.83 | 19.12 | 22.15 |

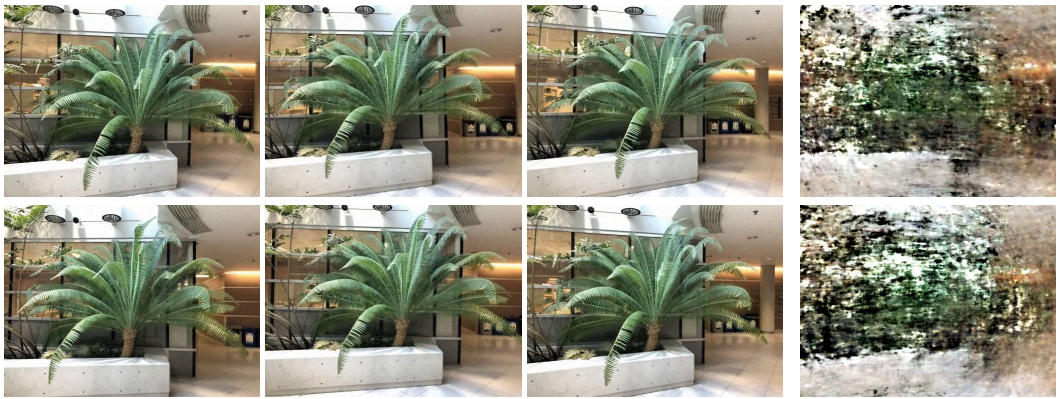

Figure 2: **Overfitting a single 3D scene.** We overfit an LFN on the context views of the "Fern" scene from NeRF [4]. Context views are reproduced almost perfectly, demonstrating that an LFN can in principle fit high-frequency detail (left). Rendering out intermediate views with unobserved rays fails, as no multi-view consistency prior is enforced (right).

two different regimes of interest. (1) The overfitting regime, where a neural scene representation is fit to a 3D scene that is completely observed, i.e., every surface point is observed enough times to enable triangulation. (2) The prior-based reconstruction regime, where we aim to reconstruct a 3D scene given incomplete observations. We note that problem (2) is of great interest in the field of artificial intelligence, where we regularly infer neural scene representations from incomplete observations. LFNs are immediately applicable to this regime, opening avenues of future work in scene understanding, scene decomposition, reinforcement learning (where a tractable inverse graphics model is of critical importance), etc.



\end{document}